\pdfoutput=1

\documentclass[11pt,a4paper]{article}
\usepackage[hyperref]{acl2018}
\usepackage{times}
\usepackage{latexsym}

\usepackage{CJK}
\usepackage{array}
\usepackage{color}

\usepackage{graphicx}
\usepackage{epstopdf}
\usepackage[font=normalsize,position=bottom]{subfig}
\usepackage{algorithm}
\usepackage[noend]{algpseudocode}
\usepackage{multirow}

\usepackage{amssymb}
\usepackage{bm}
\usepackage{balance}
\usepackage{arydshln}
\usepackage{mathtools}
\usepackage{soul}

\usepackage{booktabs}
\usepackage{amsmath,amsfonts}
\usepackage{xspace}

\usepackage{bbm}

\usepackage{xspace}
\usepackage{textcomp}

\usepackage{tabularx}

\def\argmin{\mathop{\rm argmin}}%
\def\max{\mathop{\rm max}}%

\aclfinalcopy 
\def\aclpaperid{1044} 

\title{Towards Robust Neural Machine Translation}

\author{Yong Cheng$^{\star}$, Zhaopeng Tu$^{\star}$, Fandong Meng$^{\star}$, Junjie Zhai$^{\star}$ and Yang Liu$^{\dagger}$ \\
$^{\star}$Tencent AI Lab, China \\
$^{\dagger}$State Key Laboratory of Intelligent Technology and Systems \\
Beijing National Research Center for Information Science and Technology \\
Department of Computer Science and Technology, Tsinghua University, Beijing, China \\
Beijing Advanced Innovation Center for Language Resources \\
{\tt chengyong3001@gmail.com} \\
{\tt \{zptu, fandongmeng, jasonzhai\}@tencent.com }\\ {\tt liuyang2011@tsinghua.edu.cn}}

\date{}

\begin{document}
\maketitle

\begin{abstract}
Small perturbations in the input can severely distort intermediate representations and thus impact translation quality of neural machine translation (NMT) models. 
In this paper, we propose to improve the robustness of NMT models with adversarial stability training. 
The basic idea is to make both the encoder and decoder in NMT models robust against input perturbations by enabling them to behave similarly for the original input and its perturbed counterpart.
Experimental results on Chinese-English, English-German and English-French translation tasks show that our approaches can not only achieve significant improvements over strong NMT systems but also improve the robustness of NMT models.
\end{abstract}

\section{Introduction}

Neural machine translation (NMT) models have advanced the state of the art by building a single neural network that can better learn representations~\cite{Cho:14,Sutskever:14}. 
The neural network consists of two components: an encoder network that encodes the input sentence into a sequence of distributed representations, based on which a decoder network generates the translation with an attention model~\cite{Bahdanau:15,Luong:15}.
A variety of NMT models derived from this encoder-decoder framework have further improved the performance of machine translation systems
~\cite{Gehring:17, Vaswani:17}.
NMT is capable of generalizing better to unseen text by exploiting word similarities in embeddings and capturing long-distance reordering by conditioning on larger contexts in a continuous way. 
 
\begin{table}[!t]
\centering
\begin{tabular}{|l|m{0.37\textwidth}|}
\hline
Input & tamen {\em bupa} kunnan zuochu weiqi AI. \\
\hline
Output  & They are not afraid of difficulties to make Go AI. \\
\hline
\hline
Input & tamen {\em bu{\color{red} wei}} kunnan zuochu weiqi AI. \\
\hline
Output & They are not afraid to make Go AI. \\
\hline
\end{tabular}
\caption{The non-robustness problem of neural machine translation. Replacing a Chinese word with its synonym (i.e., ``{\em bupa}'' $\rightarrow$ ``{\em buwei}'') leads to significant erroneous changes in the English translation. Both ``{\em bupa}'' and ``{\em buwei}'' can be translated to the English phrase ``{\em be not afraid of}.'' }
\label{table:example}
\end{table}

However, studies reveal that very small changes to the input can fool state-of-the-art neural networks with high probability~\cite{Goodfellow:14b,Szegedy:2014:ICML}.~\newcite{Belinkov:17} confirm this finding by pointing out that NMT models are very brittle and easily falter when presented with noisy input.  In NMT, due to the introduction of RNN and attention, each contextual word can influence the model prediction in a global context, which is analogous to the ``butterfly effect.'' As shown in Table \ref{table:example}, although we only replace a source word with its synonym, the generated translation has been completely distorted.
We investigate severe variations of translations caused by small input perturbations by replacing
one word in each sentence of a test set with its synonym. 
We observe that $69.74$\% of translations have changed and the BLEU score is only $79.01$ between the translations of the original inputs and the translations of the perturbed inputs, suggesting that NMT models are very sensitive to small perturbations in the input. The vulnerability and instability of NMT models limit their applicability to a broader range of tasks, which require robust performance on noisy inputs. For example, simultaneous translation systems use automatic speech recognition (ASR) to transcribe input speech into a sequence of hypothesized words, which are subsequently fed to a translation system. 
In this pipeline, ASR errors are presented as sentences with noisy perturbations (the same pronunciation but incorrect words), which is a significant challenge for current NMT models. Moreover, instability makes NMT models sensitive to misspellings and typos in text translation.

In this paper, we address this challenge with {\em adversarial stability training} for neural machine translation. The basic idea is to improve the robustness of two important components in NMT: the encoder and decoder. To this end, we propose two approaches to constructing noisy inputs with small perturbations to make NMT models resist them. As important intermediate representations encoded by the encoder,  they directly determine the accuracy of final translations. We introduce adversarial learning to make behaviors of the encoder consistent for both an input and its perturbed counterpart. To improve the stability of the decoder,
our method jointly maximizes the likelihoods of original and perturbed data. Adversarial stability training has the following advantages:

\begin{enumerate}
\item {\em Improving both the robustness and translation performance}: Our adversarial stability training is capable of not only improving the robustness of NMT models but also achieving better translation performance.

\item {\em Applicable to arbitrary noisy perturbations}: In this paper, we propose two approaches to constructing noisy perturbations for inputs. However, our training framework can be easily extended to arbitrary noisy perturbations. Especially, we can design task-specific perturbation methods.

\item {\em Transparent to  network architectures}: Our adversarial stability training does not depend on specific NMT architectures. It can be applied to arbitrary NMT systems. 
\end{enumerate}

Experiments on Chinese-English, English-French and English-German translation tasks show that adversarial stability training achieves significant improvements across different languages pairs. Our NMT system outperforms the state-of-the-art RNN-based NMT system (GNMT) \cite{Wu:16} and obtains comparable performance with the CNN-based NMT system \cite{Gehring:17}. Related experimental analyses validate that our training approach can improve the robustness of NMT models.

\section{Background}



NMT is an end-to-end framework which directly optimizes the translation probability of a target sentence $\mathbf{y} = y_{1},...,y_{N}$ given its corresponding source sentence $\mathbf{x} = x_{1},...,x_{M}$:
\begin{eqnarray}
P(\mathbf{y}|\mathbf{x};\bm{\theta}) = \prod_{n=1}^{N} P(y_{n}|
\mathbf{y}_{<n}, \mathbf{x}; \bm{\theta})
\end{eqnarray}
where $\bm{\theta}$ is a set of model parameters and $\mathbf{y}_{<n}$ is a partial translation.
$P(\mathbf{y}|\mathbf{x};\bm{\theta})$ is defined on a holistic neural network which mainly includes two core components: an {\em encoder} encodes a source sentence $\mathbf{x}$ into a sequence of hidden representations $\mathbf{H}_{\mathbf{x}} = \mathbf{H}_{1},...,\mathbf{H}_{M}$, and a {\em decoder} generates the $n$-th target word based on the sequence of hidden representations:
\begin{eqnarray}
P(y_{n}|\mathbf{y}_{<n}, \mathbf{x}; \bm{\theta}) \propto \exp \{ g(y_{n-1}, \mathbf{s}_{n}, \mathbf{H}_{\mathbf{x}} ;\bm{\theta}) \}
\end{eqnarray}
where $\mathbf{s}_{n}$ is the $n$-th hidden state on target side. 
Thus the model parameters of NMT include the parameter sets of the encoder $\bm{\theta}_{\rm enc}$ and the decoder $\bm{\theta}_{\rm dec}$: $\bm{\theta} = \{ \bm{\theta}_{\rm enc}, \bm{\theta}_{\rm dec} \} $.
The standard training objective is to minimize the negative log-likelihood of the training corpus $\mathcal{S} = \{\langle \mathbf{x}^{(s)}, \mathbf{y}^{(s)} \rangle \}_{s=1}^{|\mathcal{S}|}$:
\begin{eqnarray}
\hat{\bm{\theta}}&=&\argmin_{\bm{\theta}}{ \mathcal{L}(\mathbf{x}, \mathbf{y} ;\bm{\theta})} \nonumber \\
&=& \argmin_{\bm{\theta}}  \Big\{ {\sum_{\langle \mathbf{x}, \mathbf{y} \rangle \in \mathcal{S}} -\log P(\mathbf{y}|\mathbf{x};\bm{\theta}) \Big\}}
\label{eqn:mle}
\end{eqnarray}

Due to the vulnerability and instability of deep neural networks, NMT models usually suffer from a drawback: small perturbations in the input can dramatically deteriorate its translation results. \citet{Belinkov:17} point out that character-based NMT models are very brittle and easily falter when presented with noisy input. We find that word-based and subword-based NMT models also confront with this shortcoming, as shown in Table~\ref{table:example}. We argue that the distributed representations should fulfill the stability expectation, which is the underlying concept of the proposed approach.
Recent work has shown that adversarially trained models can be made robust to such perturbations~\cite{Zheng:2016:CVPR,Madry:2018:ICLR}. Inspired by this, in this work, we improve the robustness of encoder representations against noisy perturbations with adversarial learning~\cite{Goodfellow:14}.

\section{Approach}

\subsection{Overview}

\begin{figure}[!t]
\begin{center}
\includegraphics[width=0.45\textwidth]{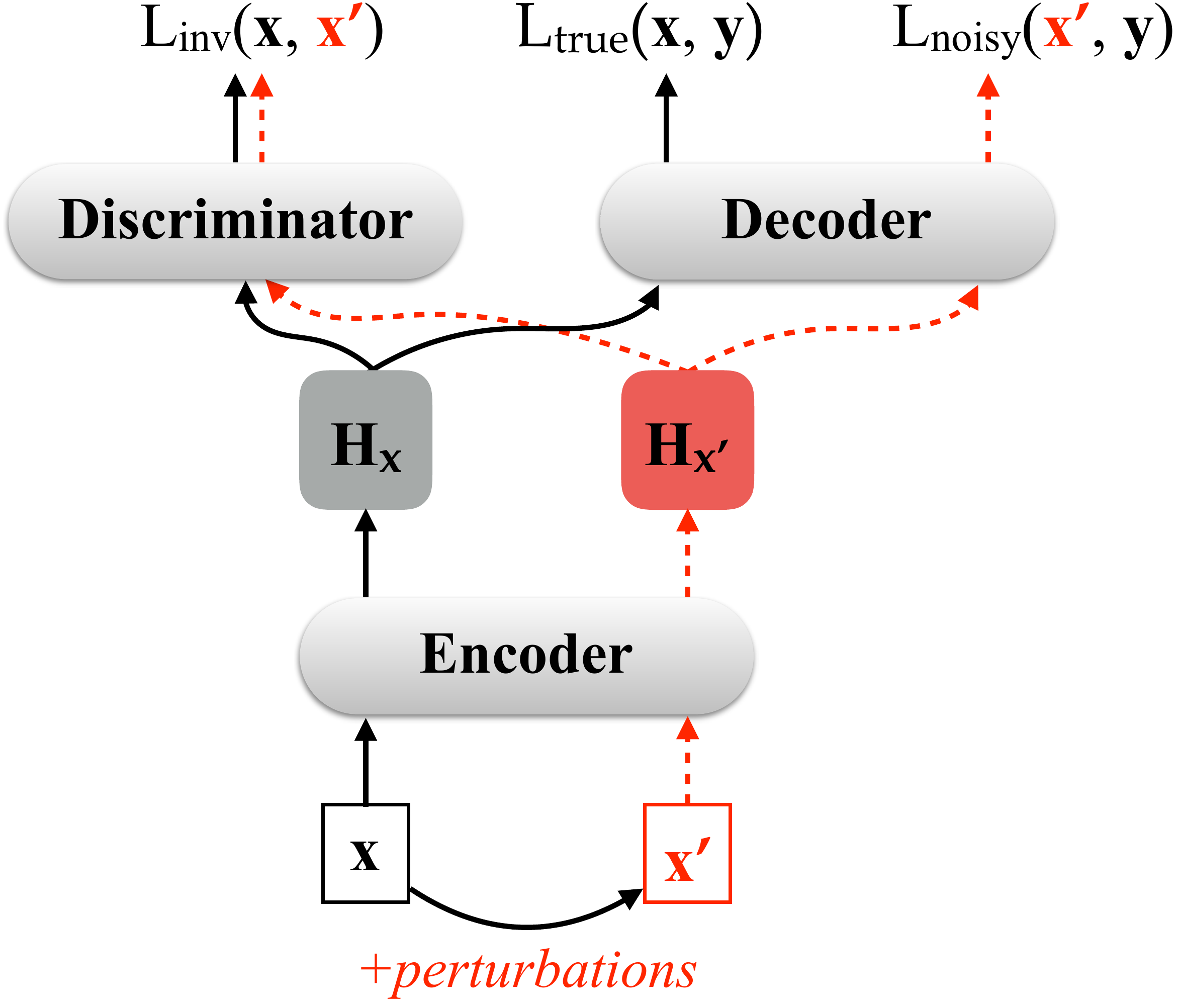}
\caption{The architecture of NMT with adversarial stability training. The dark solid arrow lines represent the forward-pass information flow for the input sentence $\mathbf{x}$, while the {\color{red}red dashed arrow lines} for the noisy input sentence $\mathbf{x}^{\prime}$, which is transformed from $\mathbf{x}$ by adding small perturbations.
} \label{fig:train} 
\end{center}
\end{figure}

The goal of this work is to propose a general approach to make NMT models learned to be more robust to input perturbations.
Our basic idea is to maintain the consistency of behaviors through the NMT model for the source sentence $\mathbf{x}$ and its perturbed counterpart $\mathbf{x}^{\prime}$.
As aforementioned, the NMT model contains two procedures for projecting a source sentence $\mathbf{x}$ to its target sentence $\mathbf{y}$:
the encoder is responsible for encoding $\mathbf{x}$ as a sequence of representations $\mathbf{H}_\mathbf{x}$, while the decoder outputs $\mathbf{y}$ with $\mathbf{H}_\mathbf{x}$ as input. 
We aim at learning the perturbation-invariant encoder and decoder.

Figure~\ref{fig:train} illustrates the architecture of our approach.
Given a source sentence $\mathbf{x}$, we construct a set of perturbed sentences $\mathcal{N}(\mathbf{x})$, in which each sentence $\mathbf{x}^{\prime}$ is constructed by adding small perturbations to $\mathbf{x}$. We require that $\mathbf{x}^{\prime}$ is a subtle variation from $\mathbf{x}$ and they have similar semantics. Given the input pair ($\mathbf{x}$, $\mathbf{x}^{\prime}$), we have two expectations: (1) the encoded representation $\mathbf{H}_{\mathbf{x}^{\prime}}$ should be close to $\mathbf{H}_\mathbf{x}$; and (2) given $\mathbf{H}_{\mathbf{x}^{\prime}}$, the decoder is able to generate the robust output $\mathbf{y}$. To this end, we introduce two additional objectives to improve the robustness of the encoder and decoder:
\begin{itemize}
    \item $\mathcal{L}_{\mathrm{inv}}(\mathbf{x}, \mathbf{x}^{\prime})$ to encourage the encoder to output similar intermediate representations $\mathbf{H}_{\mathbf{x}}$ and $\mathbf{H}_{\mathbf{x}^{\prime}}$ for $\mathbf{x}$ and $\mathbf{x}^{\prime}$ to achieve an invariant encoder, which benefits outputting the same translations. We cast this objective in the adversarial learning framework.
    \item $\mathcal{L}_{\mathrm{noisy}}(\mathbf{x}^{\prime}, \mathbf{y})$ to guide the decoder to generate output $\mathbf{y}$ given the noisy input $\mathbf{x}^{\prime}$, which is modeled as $-\log P(\mathbf{y} | \mathbf{x}^{\prime})$. It can also be defined as KL divergence between $P(\mathbf{y} | \mathbf{x})$ and $P(\mathbf{y} | \mathbf{x}^{\prime})$ that indicates using $P(\mathbf{y} | \mathbf{x})$ to teach $P(\mathbf{y} | \mathbf{x}^{\prime})$.
\end{itemize}
As seen, the two introduced objectives aim to improve the robustness of the NMT model which can be free of high variances in target outputs caused by small perturbations in inputs. It is also natural to introduce the original training objective $\mathcal{L}(\mathbf{x}, \mathbf{y})$ on $\mathbf{x}$ and $\mathbf{y}$, which can guarantee good translation performance while keeping the stability of the NMT model.

Formally, given a training corpus $\mathcal{S}$,
the adversarial stability training objective is
\begin{eqnarray}
&&\mathcal{J}(\bm{\theta}) \nonumber \\
&=& \sum_{\langle \mathbf{x}, \mathbf{y} \rangle \in \mathcal{S}} \Big(  \mathcal{L}_{\mathrm{true}}(\mathbf{x},\mathbf{y};\bm{\theta}_{\mathrm{enc}}, \bm{\theta}_{\mathrm{dec}})\quad  \nonumber \\
&& + \alpha \sum_{\mathbf{x}^{\prime} \in \mathcal{N}(\mathbf{x})} \mathcal{L}_{\mathrm{inv}}(\mathbf{x},\mathbf{x}^{\prime};\bm{\theta}_{\mathrm{enc}}, \bm{\theta}_{\mathrm{dis}}) \quad \nonumber \\
&& + \beta \sum_{\mathbf{x}^{\prime} \in \mathcal{N}(\mathbf{x})} \mathcal{L}_{\mathrm{noisy}}(\mathbf{x}^{\prime},\mathbf{y};\bm{\theta}_{\mathrm{enc}}, \bm{\theta}_{\mathrm{dec}})\Big)
\label{eq:joint_training}
\end{eqnarray} 
where $\mathcal{L}_{\mathrm{true}}(\mathbf{x}, \mathbf{y})$ and $\mathcal{L}_{\mathrm{noisy}}(\mathbf{x}^{\prime}, \mathbf{y})$ are calculated using Equation~\ref{eqn:mle}, and $\mathcal{L}_{\mathrm{inv}}(\mathbf{x},\mathbf{x}^{\prime})$ is the adversarial loss to be described in Section~\ref{sec-encoder}.
$\alpha$ and $\beta$ control the balance between the original translation task and the stability of the NMT model. 
$\bm{\theta} = \{\bm{\theta}_{\rm enc}, \bm{\theta}_{\rm dec},\bm{\theta}_{\rm dis}\}$ are trainable parameters of the encoder, decoder, and the newly introduced discriminator used in adversarial learning. 
As seen, the parameters of encoder $\bm{\theta}_{\rm enc}$ and decoder $\bm{\theta}_{\rm dec}$ are trained to minimize both the translation loss $\mathcal{L}_{\mathrm{true}}(\mathbf{x}, \mathbf{y})$ and the stability losses ($\mathcal{L}_{\mathrm{noisy}}(\mathbf{x}^{\prime}, \mathbf{y})$ and $\mathcal{L}_{\mathrm{inv}}(\mathbf{x},\mathbf{x}^{\prime})$). 

Since $\mathcal{L}_{\mathrm{noisy}}(\mathbf{x}^{\prime}, \mathbf{y})$ evaluates the translation loss on the perturbed neighbour $\mathbf{x}^{\prime}$ and its corresponding target sentence $\mathbf{y}$, it means that we augment the training data by adding perturbed neighbours, which can potentially improve the translation performance.
In this way, our approach not only makes the output of NMT models more robust, but also improves the performance on the original translation task.

In the following sections, we will first describe how to construct perturbed inputs with different strategies to fulfill different goals (Section~\ref{sec-perturbed-input}), followed by the proposed adversarial learning mechanism for the perturbation-invariant encoder (Section~\ref{sec-encoder}). We conclude this section with the training strategy (Section~\ref{sec-train}).

\subsection{Constructing Perturbed Inputs}
\label{sec-perturbed-input}

At each training step, we need to generate a perturbed neighbour set $\mathcal{N}(\mathbf{x})$ for each source sentence $\mathbf{x}$ for adversarial stability training. 
In this paper, we propose two strategies to construct the perturbed inputs at multiple levels of representations.  

The first approach generates perturbed neighbours at the {\em lexical} level. Given an input sentence $\mathbf{x}$, we randomly sample some word positions to be modified. Then we replace words at these positions with other words in the vocabulary according to the following distribution:
\begin{eqnarray}
P(x|\mathbf{x}_{i})=\frac{\exp{\left \{ \cos{(\mathbf{E}[\mathbf{x}_{i}], \mathbf{E}[x])}\right\}}}{\sum_{x \in \mathcal{V}_{x} \backslash\mathbf{x}_{i}}\exp{\left \{\cos{(\mathbf{E}[\mathbf{x}_{i}], \mathbf{E}[x])}\right\}}}
\end{eqnarray}
where $\mathbf{E}[\mathbf{x}_{i}]$ is the word embedding for word $\mathbf{x}_{i}$, $\mathcal{V}_{x}\backslash \mathbf{x}_{i}$ is the source vocabulary set excluding the word $\mathbf{x}_{i}$, and $\cos{(\mathbf{E}[\mathbf{x}_{i}], \mathbf{E}[x])}$ measures the similarity between word $\mathbf{x}_{i}$ and $x$. Thus we can change the word to another word with similar semantics.

One potential problem of the above strategy is that it is hard to enumerate all possible positions and possible types to generate perturbed neighbours.
Therefore, we propose a more general approach to modifying the sentence at the {\em feature} level. Given a sentence, we can obtain the word embedding for each word. We add the Gaussian noise to a word embedding to simulate possible types of perturbations. That is
\begin{eqnarray}
\mathbf{E}[\mathbf{x}^{\prime}_{i}] = \mathbf{E}[\mathbf{x}_{i}] + \bm{\epsilon}, \quad
\bm{\epsilon} \sim \mathbf{N}(0, \sigma^{2}\mathbf{I}) \label{eqn:gaussian}
\end{eqnarray}
where the vector $\bm{\epsilon}$ is sampled from a Gaussian distribution with variance $\sigma^{2}$. $\sigma$ is a hyper-parameter. We simply introduce Gaussian noise to all of word embeddings in $\mathbf{x}$.

The proposed scheme is a general framework where one can freely define the strategies to construct perturbed inputs.
We just present two possible examples here. The first strategy is potentially useful when the training data contains noisy words, while the latter is a more general strategy to improve the robustness of common NMT models.
In practice, one can design specific strategies for particular tasks. For example, we can replace correct words with their homonyms (same pronunciation but different meanings) to improve NMT models for simultaneous translation systems.
 

\subsection{Adversarial Learning for the Perturbation-invariant Encoder}
\label{sec-encoder}

The goal of the perturbation-invariant encoder is to make the representations produced by the encoder indistinguishable when fed with a correct sentence $\mathbf{x}$ and its perturbed counterpart $\mathbf{x}^{\prime}$, which is directly beneficial to the output robustness of the decoder. We cast the problem in the adversarial learning framework~\cite{Goodfellow:14}.
The encoder serves as the generator $G$, which defines the policy that generates a sequence of hidden representations $\mathbf{H}_{\mathbf{x}}$ given an input sentence $\mathbf{x}$.
We introduce an additional discriminator $D$ to distinguish the representation of perturbed input $\mathbf{H}_{\mathbf{x}^{\prime}}$ from that of the original input $\mathbf{H}_{\mathbf{x}}$.
The goal of the generator $G$ (i.e., encoder) is to produce similar representations for $\mathbf{x}$ and $\mathbf{x}^{\prime}$ which could fool the discriminator, while the discriminator $D$ tries to correctly distinguish the two representations.

Formally, the adversarial learning objective is
\begin{eqnarray}
&&\mathcal{L}_{\mathrm{inv}}(\mathbf{x}, \mathbf{x}^{\prime};\bm{\theta}_{\mathrm{enc}}, \bm{\theta}_{\mathrm{dis}}) \nonumber \\
&=& \mathbb{E}_{\mathbf{x} \sim \mathcal{S}}[-\log D(G(\mathbf{x}))] + \nonumber \\
&& \mathbb{E}_{\mathbf{x}^{\prime} \sim \mathcal{N}(\mathbf{x})} \left[ - \log (1-D(G(\mathbf{x}^{\prime})))\right]   
\label{eq:adv}
\end{eqnarray}
The discriminator outputs a classification score given an input representation, and tries to maximize $D(G(\mathbf{x}))$ to 1 and minimize $D(G(\mathbf{x}^{\prime}))$ to 0.
The objective encourages the encoder to output similar representations for $\mathbf{x}$ and $\mathbf{x}^{\prime}$, so that the discriminator fails to distinguish them. 

The training procedure can be regarded as a min-max two-player game.
The encoder parameters $\bm{\theta}_{\rm enc}$ are trained to maximize the loss function to fool the discriminator. The discriminator parameters $\bm{\theta}_{\rm dis}$ are optimized to minimize this loss for improving the discriminating ability. For efficiency, we update both the encoder and the discriminator simultaneously at each iteration, rather than the periodical training strategy that is commonly used in adversarial learning. \citet{Lamb:16} also propose a similar idea to use Professor Forcing to make the behaviors of RNNs be indistinguishable when training and sampling the networks.

\subsection{Training}
\label{sec-train}

As shown in Figure \ref{fig:train}, our training objective includes three sets of model parameters for three modules. We use mini-batch stochastic gradient descent to optimize our model. In the forward pass, besides a mini-batch of $\mathbf{x}$ and $\mathbf{y}$, we also construct a mini-batch consisting of the perturbed neighbour $\mathbf{x}^{\prime}$ and $\mathbf{y}$. We propagate the information to calculate these three loss functions according to arrows. Then, gradients are collected to update three sets of model parameters. Except for the gradients of $\mathcal{L}_{\rm inv}$ with respect to $\bm{\theta}_{\rm enc}$ are multiplying by $-1$, other gradients are normally back-propagated. Note that we update $\bm{\theta}_{\rm inv}$ and $\bm{\theta}_{\rm enc}$ simultaneously for training efficiency.
\begin{table*}[!t]
\centering
\begin{tabular}{l|l|l|lllll}
{System} &{Training} &{MT06} &{MT02} & {MT03} & {MT04} & {MT05} &{MT08}\\
\hline\hline
\citet{Shen:15} &MRT &37.34 &40.36 &40.93 &41.37 &38.81 &29.23  \\
\citet{Wang:17} &MLE &37.29 &-- &39.35 &41.15 &38.07   & --   \\
\citet{Zhang:2018:AAAI} &   MLE &  38.38  &  --  &  40.02  & 42.32 &  38.84 & -- \\
\hline \hline
\multirow{3}{*}{{\em this work}}&MLE &41.38 &43.52 &41.50 &43.64 &41.58  &31.60 \\
&AST$_{\rm lexical}$ &43.57 &44.82 &42.95 &45.05 &43.45 &  34.85  \\
&AST$_{\rm feature}$  & \bf 44.44 & \bf 46.10  & \bf 44.07 & \bf 45.61 &  \bf 44.06 & \bf 34.94 \\
\end{tabular}
\caption{Case-insensitive BLEU scores on Chinese-English translation.} 
\label{table:comparison_zhen}
\end{table*}

\begin{table*}[!t]
\centering
\begin{tabular}{l|l|l|l}
System &Architecture & Training &BLEU \\
\hline \hline
\citet{Shen:15} & Gated RNN with 1 layer &MRT  &20.45 \\
\citet{Luong:15} &LSTM with 4 layers &MLE &20.90 \\
\citet{Kalchbrenner:16} & ByteNet with 30 layers &MLE &23.75\\
\citet{Wang:17} &DeepLAU with 4 layers &MLE & 23.80\\
\citet{Wu:16} &LSTM with 8 layers &RL & 24.60 \\
\citet{Gehring:17} & CNN with 15 layers &MLE &25.16 \\
\citet{Vaswani:17} & Self-attention with 6 layers &MLE & 28.40 \\
\hline
\hline
\multirow{3}{*}{{\em this work}} & {\multirow{3}{*}{Gated RNN with 2 layers}} &MLE & 24.06 \\
& &AST$_{\rm lexical}$ &25.17\\
& &AST$_{\rm feature}$ & \bf25.26\\
\end{tabular}
\caption{Case-sensitive BLEU scores on WMT 14 English-German translation.}
\label{table:comparison_ende}
\end{table*}

\begin{table}[!t]
\centering
\begin{tabular}{l|ll}
Training &tst2014 &tst2015 \\
\hline\hline
MLE &36.92 & 36.90\\
AST$_{\rm lexical}$ &37.35 &37.03 \\
AST$_{\rm feature}$ & \bf38.03 &\bf37.64 \\
\end{tabular}
\caption{Case-sensitive BLEU scores on IWSLT English-French translation.}
\label{table:comparison_enfr}
\end{table}

\section{Experiments}

\subsection{Setup}
We evaluated our adversarial stability training on translation tasks of several language pairs, and reported the 4-gram BLEU \cite{Papineni:02} score as calculated by the {\em multi-bleu.perl} script.

\noindent{\bf Chinese-English}
We used the LDC corpus consisting of 1.25M sentence pairs with 27.9M Chinese words
and 34.5M English words respectively.  We selected the best model using the NIST 2006 set as the validation set (hyper-parameter optimization and model selection). The NIST 2002, 2003, 2004, 2005, and
2008 datasets are used as test sets.

\noindent{\bf English-German}
We used the WMT 14 corpus containing 4.5M sentence pairs with 118M English words and 111M German words. The validation set is newstest2013, and the test set is newstest2014. 

\noindent{\bf English-French}
We used the IWSLT corpus which contains 0.22M sentence pairs with 4.03M English words and 4.12M French words. The IWLST corpus is
very dissimilar from the NIST and WMT corpora. As they are collected from TED talks and inclined to spoken language, we want to verify our approaches on the non-normative text. The IWSLT 14 test set is taken as the validation set and 15 test set is used as the test set.

For English-German and English-French, we tokenize both English, German and French words using \verb|tokenize.perl| script. We follow \citet{Sennrich:15a} to split words into subword units. The numbers of merge operations in byte pair encoding (BPE) are set to 30K, 40K and 30K respectively for Chinese-English, English-German, and English-French. We report the case-sensitive tokenized BLEU score for English-German and English-French and the case-insensitive tokenized BLEU score for Chinese-English.

Our baseline system is an in-house NMT system. Following \citet{Bahdanau:15}, we implement an RNN-based NMT in which both the encoder and decoder are two-layer RNNs with residual connections between layers \cite{He:16}. The gating mechanism of RNNs is gated recurrent unit (GRUs) \cite{Cho:14}. We apply layer normalization \cite{Ba:16} and dropout \cite{Hinton:12} to the hidden states of GRUs. Dropout is also added to the source and target word embeddings. We share the same matrix between the target word embeedings and the pre-softmax linear transformation \cite{Vaswani:17}. We update the set of model parameters using Adam SGD \cite{Kingma:14}. Its learning rate is initially set to $0.05$ and varies according to the formula in \citet{Vaswani:17}.

Our adversarial stability training initializes the model based on the parameters trained by maximum likelihood estimation (MLE). We denote adversarial stability training based on
lexical-level perturbations and feature-level perturbations respectively as AST$_{\rm lexical}$ and AST$_{\rm feature}$. We only sample one perturbed neighbour $\mathbf{x}^{\prime} \in \mathcal{N}(\mathbf{x})$ for training efficiency. For the discriminator used in $\mathcal{L}_{\mathrm{inv}}$, we adopt the CNN discriminator proposed by \citet{Kim:14} to address the variable-length problem of the sequence generated by the encoder. In the CNN discriminator, the filter windows are set to 3, 4, 5 and rectified linear units are applied after convolution operations.  We tune the hyper-parameters on the validation set through a grid search. We find that both the optimal values of $\alpha$ and $\beta$ are set to $1.0$. The standard variance in Gaussian noise used in the formula (\ref{eqn:gaussian}) is set to $0.01$. The number of words that are replaced in the sentence $\mathbf{x}$ during lexical-level perturbations is taken as $\max(0.2|\mathbf{x}|, 1)$ in which $|\mathbf{x}|$ is the length of $\mathbf{x}$. The default beam size for decoding is $10$.

\subsection{Translation Results}

\subsubsection{NIST Chinese-English Translation}

Table \ref{table:comparison_zhen} shows the results on Chinese-English translation. Our strong baseline system significantly outperforms
previously reported results on Chinese-English NIST datasets trained on RNN-based NMT.  \citet{Shen:15} propose minimum risk training (MRT) for NMT, which directly optimizes model parameters with respect to BLEU scores. \citet{Wang:17} address the issue of severe gradient diffusion with linear associative units (LAU). Their system is deep with an encoder of 4 layers and a decoder of 4 layers.
\citet{Zhang:2018:AAAI} propose to exploit both left-to-right and right-to-left decoding strategies for NMT to capture bidirectional dependencies. Compared with them, our NMT system trained by MLE outperforms their best models by around 3 BLEU points. We hope that the strong baseline systems used in this work make the evaluation convincing.

We find that introducing adversarial stability training into NMT can bring substantial improvements over previous work (up to $+3.16$ BLEU points over \citet{Shen:15},  up to $+3.51$ BLEU points over \citet{Wang:17} and up to $+2.74$ BLEU points over \citet{Zhang:2018:AAAI}) and our system trained with MLE across all the datasets. Compared with our baseline system, AST$_{\rm lexical}$ achieves $+1.75$ BLEU improvement on average. AST$_{\rm feature}$ performs better, which can obtain $+2.59$ BLEU points on average and up to $+3.34$ BLEU points on NIST08.

\begin{table*}[!t]
\centering
\begin{tabular}{l|l|llllll}
{Synthetic Type}& {Training}  &{0 Op.} &{1 Op.} & {2 Op.} & {3 Op.} & {4 Op.} &{5 Op.} \\
\hline
\hline 
\multirow{3}{*}{Swap} &MLE &41.38 &38.86 &37.23 &35.97 &34.61 &32.96 \\
&AST$_{\rm lexical}$ &43.57 &41.18 &39.88 &37.95 &37.02 &36.16 \\
&AST$_{\rm feature}$ &44.44 &42.08 &40.20 &38.67 &36.89 &35.81 \\
\hline
\hline
\multirow{3}{*}{Replacement} &MLE &41.38 &37.21 &31.40 &27.43 &23.94 &21.03 \\
&AST$_{\rm lexical}$ &43.57 &40.53 &37.59 &35.19 &32.56 &30.42 \\
&AST$_{\rm feature}$ &44.44 &40.04 &35.00 &30.54 &27.42 &24.57 \\\hline
\hline
\multirow{3}{*}{Deletion} &MLE &41.38 &38.45 &36.15 &33.28 &31.17 &28.65 \\
&AST$_{\rm lexical}$ &43.57 &41.89 &38.56 &36.14 &34.09 &31.77 \\
&AST$_{\rm feature}$ &44.44 &41.75 &39.06 &36.16 &33.49 &30.90 \

\end{tabular}
\caption{Translation results of synthetic perturbations on the validation set in Chinese-English translation. 
``1 Op.'' denotes that we conduct one operation (swap, replacement or deletion) on the original sentence.}
\label{table:comparison_synthetic}
\end{table*}

\begin{table*}[!t]
\centering
\begin{tabular}{|l|l|}
\hline
Source & zhongguo dianzi yinhang yewu guanli xingui jiangyu sanyue yiri qi shixing \\
\hline
Reference & china's new management rules for e-banking operations to take effect on march 1 \\
\hline
\hline 
MLE & china's electronic bank rules to be implemented on march 1 \\
\hline
\multirow{2}{*}{AST$_{\rm lexical}$} &new rules for business administration of china 's electronic banking industry \\
&will come into effect on march 1 . \\
\hline 
\multirow{2}{*}{AST$_{\rm feature}$} &new rules for business management of china 's electronic banking industry to\\
&come into effect on march 1 \\
\hline \hline
Perturbed Source & {\em zhong{\color{red} fang}} dianzi yinhang yewu guanli xingui jiangyu sanyue yiri qi shixing \\
\hline
MLE & china to implement new regulations on business management \\
\hline
\multirow{2}{*}{AST$_{\rm lexical}$} & the new regulations for the business administrations of the chinese electronics \\
			       &	 bank will come into effect on march 1 .\\
\hline	       
\multirow{2}{*}{AST$_{\rm feature}$} & new rules for business management of china's electronic banking industry to \\
			       &	come into effect on march 1 \\
\hline
\end{tabular}
\caption{Example translations of a source sentence and its perturbed counterpart by replacing a Chinese word ``zhongguo'' with its synonym ``zhongfang.''} \label{table:trans_example}
\end{table*}

\subsubsection{WMT 14 English-German Translation}
In Table \ref{table:comparison_ende}, we list existing NMT systems as comparisons.  All these systems use the same WMT 14 English-German corpus. Except that
\citet{Shen:15} and \citet{Wu:16} respectively adopt MRT and reinforcement learning (RL), other systems all use MLE as training criterion. All the
systems except for \citet{Shen:15} are deep NMT models with no less than four layers. Google's neural machine translation (GNMT) \cite{Wu:16} represents a
strong RNN-based NMT system. Compared with other RNN-based NMT systems except for GNMT, our baseline system with two layers can achieve better performance than theirs.

When training our NMT system with AST$_{\rm leixcal}$, significant improvement ($+1.11$ BLEU points) can be observed. AST$_{\rm feature}$ can obtain slightly better performance. 
Our NMT system outperforms the state-of-the-art RNN-based NMT system, GNMT, with $+0.66$ BLEU point and performs comparably with 
\citet{Gehring:17} which is based on CNN with 15 layers. 
Given that our approach can be applied to any NMT systems, we expect that the adversarial stability training mechanism can further improve performance upon the advanced NMT architectures. We leave this for future work.

\subsubsection{IWSLT English-French Translation}
Table \ref{table:comparison_enfr} shows the results on IWSLT English-French Translation. Compared with our  strong baseline system trained by MLE, we observe that our models consistently improve translation performance in all datasets.
AST$_{\rm feature}$ can achieve significant improvements on the  tst2015 although AST$_{\rm lexical}$ obtains comparable results. These
demonstrate that our approach maintains good performance on the non-normative text.

\subsection{Results on Synthetic Perturbed Data}

In order to investigate the ability of our training approaches to deal with perturbations, we experiment with three types of synthetic perturbations:
\begin{itemize}
  \setlength{\itemsep}{5pt}
  \setlength{\parskip}{0pt}
  \setlength{\parsep}{0pt}
    \item \textbf{Swap}: We randomly choose $N$ positions from a sentence and then swap the chosen words with their right neighbours.
    \item \textbf{Replacement}: We randomly replace sampled words in the sentence with other words.
    \item \textbf{Deletion}: We randomly delete $N$ words from each sentence in the dataset.
\end{itemize}

As shown in Table \ref{table:comparison_synthetic},  we can find that our training approaches, AST$_{\rm lexical}$ and AST$_{\rm feature}$, consistently outperform MLE against perturbations on all the numbers of operations.  This means that our approaches have the capability of resisting perturbations. Along with the number of operations increasing, the performance on MLE drops quickly. Although the performance of  our approaches also drops, we can see that our approaches consistently surpass MLE. In AST$_{\rm lexical}$, with 0 operation, the difference is +2.19 (43.57 Vs. 41.38) for all synthetic types, but the differences are enlarged to +3.20, +9.39, and +3.12 respectively for the three types with 5 operations. 

In the {\em Swap} and {\em Deletion} types,  AST$_{\rm lexical}$ and AST$_{\rm feature}$ perform comparably after more than four operations. Interestingly, AST$_{\rm lexical}$ performs significantly better than both of MLE and AST$_{\rm feature}$ after more than one operation in the {\em Replacement} type. This is because AST$_{\rm lexical}$ trains the model specifically on perturbation data that is constructed by replacing words, which agrees with the {\em Replacement} Type. Overall, AST$_{\rm lexical}$ performs better than AST$_{\rm feature}$ against perturbations after multiple operations. We speculate that the perturbation method for AST$_{\rm lexical}$ and synthetic type are both discrete and they keep more consistent. Table \ref{table:trans_example} shows example translations of a Chinese sentence and its perturbed counterpart.

These findings indicate that we can construct specific perturbations for a particular task. For example, in simultaneous translation, an automatic speech recognition system usually generates wrong words with the same pronunciation of correct words, which dramatically affects the quality of  machine translation system. Therefore, we can design specific perturbations aiming for this task.

\subsection{Analysis}
\begin{table}[!t]
\centering
\begin{tabular}{ccc|l}
$\mathcal{L}_{\rm true}$ & $\mathcal{L}_{\rm noisy}$ &$\mathcal{L}_{\rm inv}$ &BLEU \\
\hline\hline
$\surd$  &$\times$ &$\times$ & 41.38\\
$\surd$  &$\times$ &$\surd$  & 41.91\\
$\times$ &$\surd$  &$\times$ & 42.20\\
$\surd$  &$\surd$  &$\times$ & 42.93\\
$\surd$  &$\surd$  &$\surd$  & 43.57\\
\end{tabular}
\caption{Ablation study of adversarial stability training AST$_{\rm lexical}$ on Chinese-English translation. ``$\surd$'' means the loss function is included in the training objective while ``$\times$'' means it is not.}
\label{table:ablation_study}
\end{table}

\begin{figure}[!t]
\begin{center}
\includegraphics[width=0.49\textwidth]{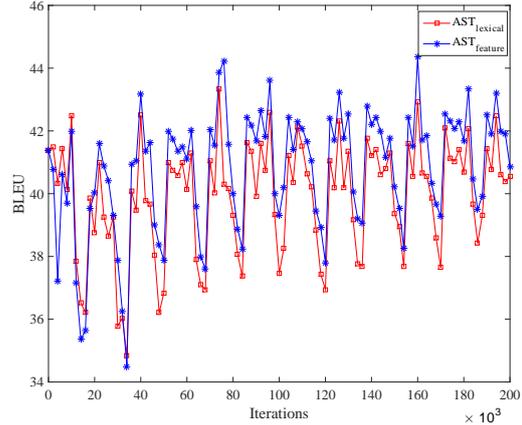}
\caption{BLEU scores of AST$_{\rm lexical}$ over iterations on Chinese-English validation set.}
\label{fig:adv_lexical_bleu} 
\end{center}
\end{figure}

\begin{figure}[!t]
\begin{center}
\includegraphics[width=0.49\textwidth]{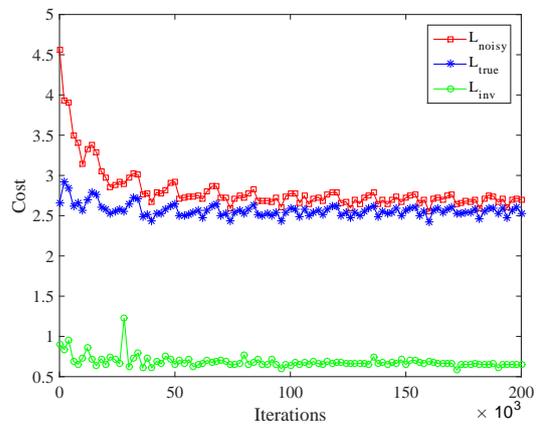}
\caption{Learning curves of three loss functions, $\mathcal{L}_{\mathrm{true}}$, $\mathcal{L}_{\mathrm{inv}}$ and $\mathcal{L}_{\mathrm{noisy}}$ over iterations on Chinese-English validation set. } \label{fig:train_cost} 
\end{center}
\end{figure}

\subsubsection{Ablation Study}
Our training objective function Eq. (\ref{eq:joint_training}) contains three loss functions. We perform an ablation study on the Chinese-English translation to understand the importance of these loss functions by choosing AST$_{\rm lexical}$ as an example. As Table \ref{table:ablation_study} shows, if we remove $\mathcal{L}_{\rm adv}$, the translation performance decreases by $0.64$ BLEU point. However, when $\mathcal{L}_{\rm noisy}$ is excluded from the training objective function, it results in a significant drop of $1.66$ BLEU point. Surprisingly, only using $\mathcal{L}_{\rm noisy}$ is able to lead to an increase of $0.88$ BLEU point.

\subsubsection{BLEU Scores over Iterations}

Figure \ref{fig:adv_lexical_bleu} shows the changes of BLEU scores over iterations respectively for AST$_{\rm lexical}$ and
AST$_{\rm feature}$. They behave nearly consistently. Initialized by the model trained by MLE, their performance drops rapidly. Then it starts to go up quickly. Compared
with the starting point, the maximal dropping points reach up to about $7.0$ BLEU points. Basically, the curves
present the state of oscillation. We think that introducing random perturbations and adversarial learning can make the training not very stable like MLE. 

\subsubsection{Learning Curves of Loss Functions}

Figure \ref{fig:train_cost} shows the learning curves of three loss functions, $\mathcal{L}_{\rm true}$, $\mathcal{L}_{\rm inv}$ and $\mathcal{L}_{\rm noisy}$. 
We can find that their costs of loss functions decrease not steadily. Similar to the Figure \ref{fig:adv_lexical_bleu}, there still exist oscillations in the learning curves although they do not change much sharply. We find that $\mathcal{L}_{\mathrm{inv}}$ converges to around $0.68$ after about $100K$ iterations, which indicates that discriminator outputs probability $0.5$ for  both positive and negative samples and it cannot distinguish them. Thus the behaviors of the encoder for $\mathbf{x}$ and its perturbed neighbour $\mathbf{x}^{\prime}$ perform nearly consistently.

\section{Related Work}
Our work is inspired by two lines of research: (1) adversarial learning  and (2) data augmentation.

\paragraph{Adversarial Learning}
Generative Adversarial Network (GAN) \cite{Goodfellow:14} and its related derivative have been widely applied in computer vision~\cite{radford2015unsupervised,salimans2016improved} and natural language processing~\cite{Li:17,Yang:2017:arXiv}. Previous work has constructed
adversarial examples to attack trained networks and make networks resist them, which has proved to improve the robustness of networks \cite{Goodfellow:14b, Miyato:15, Zheng:2016:CVPR}. \citet{Belinkov:17} introduce adversarial examples to training data for character-based NMT models. In contrast to theirs, adversarial stability training aims to stabilize both the encoder and decoder in NMT models. We adopt adversarial learning to learn the perturbation-invariant encoder.

\paragraph{Data Augmentation}
Data augmentation has the capability to improve the robustness of NMT models. In NMT, there is a number of work that augments the training data with monolingual corpora \cite{Sennrich:15, Cheng:16, Hedual:16, Zhang:16}. They all leverage complex models such as inverse NMT models to generate translation equivalents for monolingual corpora. Then they augment the parallel corpora with these pseudo corpora to improve NMT models. Some authors have recently endeavored to achieve zero-shot NMT through transferring knowledge from bilingual corpora of other language pairs \cite{Chen:17,Zheng:17,Cheng:17} or monolingual corpora \cite{Lample:18,Artetxe:18}. Our work significantly differs from these work. We do not resort to any complicated 
models to generate perturbed data and do not depend on extra monolingual or bilingual corpora. The way we exploit is more convenient and easy to implement. We focus more on improving the robustness of NMT models.

\section{Conclusion}

We have proposed adversarial stability training to improve the robustness of NMT models.
The basic idea is to train both the encoder and decoder robust to input perturbations by enabling them to behave similarly  for  the  original  input  and  its  perturbed  counterpart. We propose two approaches to construct perturbed data to adversarially train the encoder and stabilize the decoder. 
Experiments on Chinese-English, English-German and English-French translation tasks show that the proposed approach can improve both the robustness and translation performance.

As our training framework is not limited to specific perturbation types, it is interesting to evaluate our approach in natural noise existing in practical applications, such as homonym in the simultaneous translation system. It is also necessary to further validate our approach on more advanced NMT architectures, such as CNN-based NMT \cite{Gehring:17} and Transformer \cite{Vaswani:17}.

\section*{Acknowledgments}
We thank the anonymous reviewers for
their insightful comments and suggestions. We also thank Xiaoling Li for analyzing experimental results and providing valuable examples. Yang Liu is supported by the National Key R\&D Program of China (No. 2017YFB0202204), National Natural Science Foundation of China (No. 61761166008, No. 61522204), Beijing Advanced Innovation Center for Language Resources, and the NExT++ project supported by the National Research Foundation, Prime Minister’s Office, Singapore under its IRC@Singapore Funding Initiative.
\balance
\bibliographystyle{acl_natbib}
\bibliography{acl2018}
\end{document}